\documentclass[runningheads]{llncs}
\usepackage[T1]{fontenc}
\usepackage{graphicx}
\usepackage{amssymb}
\usepackage{amsmath}
\usepackage{bbm}
\usepackage{bm}
\usepackage{multirow}
\usepackage{float}
\usepackage{xcolor}
\usepackage{marvosym}
\usepackage{orcidlink}

\begin{document}
\title{GeoT: Geometry-guided Instance-dependent Transition Matrix for Semi-supervised \\Tooth Point Cloud Segmentation}
\titlerunning{GeoT}

\author{Weihao Yu\inst{1}~\orcidlink{0000-0003-1718-1859} \and
Xiaoqing Guo\inst{2,3}~\orcidlink{0000-0002-9476-521X} \and
Chenxin Li\inst{1}~\orcidlink{0000-0002-8341-4801} \and 
Yifan Liu\inst{1}~\orcidlink{0000-0003-2887-7704} \and \\
Yixuan Yuan\inst{1}\textsuperscript{(\Letter)}~\orcidlink{0000-0002-0853-6948}}   
%
\authorrunning{YU et al.}
%
\institute{Department of Electronic Engineering, \\ 
The Chinese University of Hong Kong, Hong Kong, China \\
\email{yxyuan@ee.cuhk.edu.hk} \and  
Department of Computer Science, \\
Hong Kong Baptist University, Hong Kong, China \and
Department of Engineering Science, University of Oxford, Oxford, UK
}
\maketitle              
\begin{abstract}
Achieving meticulous segmentation of tooth point clouds from intra-oral scans stands as an indispensable prerequisite for various orthodontic applications. Given the labor-intensive nature of dental annotation, a significant amount of data remains unlabeled, driving increasing interest in semi-supervised approaches. One primary challenge of existing semi-supervised medical segmentation methods lies in noisy pseudo labels generated for unlabeled data. To address this challenge, we propose GeoT, the first framework that employs instance-dependent transition matrix (IDTM) to explicitly model noise in pseudo labels for semi-supervised dental segmentation. Specifically, to handle the extensive solution space of IDTM arising from tens of thousands of dental points, we introduce tooth geometric priors through two key components: point-level geometric regularization (PLGR) to enhance consistency between point adjacency relationships in 3D and IDTM spaces, and class-level geometric smoothing (CLGS) to leverage the fixed spatial distribution of tooth categories for optimal IDTM estimation. Extensive experiments performed on the public Teeth3DS dataset and private dataset demonstrate that our method can make full utilization of unlabeled data to facilitate segmentation, achieving performance comparable to fully supervised methods with only $20\%$ of the labeled data. Code will be public available at \url{https://github.com/CUHK-AIM-Group/GeoT}
\keywords{Semi-supervised learning  \and Point cloud segmentation \and Geometric prior.}
\end{abstract}
\section{Introduction}
Computer-aided design (CAD) systems have facilitated the widespread adoption of intra-oral scans (IOS) in essential tasks such as dental restoration, treatment planning, and orthodontic diagnosis~\cite{almukhtar2014comparison,hajeer2004current}. Accurate tooth point cloud segmentation is crucial for these applications~\cite{im2022accuracy,liu2022hierarchical}. However, manual annotation of IOS point clouds is prohibitively labor-intensive, driving research toward automated solutions. Fully-supervised methods~\cite{cui2021tsegnet,hao2022toward,lian2020deep,xu20183d,zanjani2019mask} have made significant progress through strategies like multi-scale local-global feature integration~\cite{lian2020deep} and confidence-aware cascading~\cite{cui2021tsegnet}, but they rely heavily on large-scale labeled data, which remains scarce due to the expertise required for dental annotations.

Recently, semi-supervised methods~\cite{chen2023decoupled,liu2022semi,xie2023deep} have garnered increased attention as they exploit unlabeled data by leveraging pseudo labels. The main challenge lies in noise within these pseudo labels. Many existing approaches employ confidence thresholds to filter unreliable data~\cite{sohn2020fixmatch,zhang2021flexmatch}. In order to fully utilize unlabeled data, the noise transition matrix (NTM) offers an alternative by modeling the label noise distribution and integrating all available pseudo labels~\cite{li2021provably,xia2019anchor}. Specifically, NTM estimates a matrix where each entry represents the probability that a sample from the true class is mislabeled as another class. By multiplying NTM with the learning objective, the network can be trained on noisy labels while accounting for their uncertainty. It preserves all available training signals rather than enforcing a hard threshold that may eliminate informative samples. Some works~\cite{cheng2022instance,cheng2020learning,xia2020part} have further extended this concept to instance-dependent transition matrix (IDTM)\footnote{For convenience, unless otherwise specified, NTM in the following discussions will refer to the instance-dependent transition matrix.}, which capture how noise patterns vary across different samples, providing a more flexible and accurate model of label noise in real-world scenarios.

Motivated by this, in this paper, we make the first attempt to utilize IDTM to handle pseudo label noise for tooth point cloud segmentation. However, it is not trivial to extend IDTM to such a segmentation task. Firstly, estimating IDTM requires appropriate assumptions to constrain the degrees of freedom of its solution space~\cite{cheng2020learning,xia2020part}. 
For instance, in image classification tasks, Xia \emph{et al.}~\cite{xia2020part} assumed that the IDTM can be decomposed into a series of partially correlated basic transition matrices, and then reconstructed the IDTM through a convex combination of these matrices. But this approach is not applicable to dense tasks such as dental segmentation, where the instance is a point and cannot be further decomposed. Secondly, each point possesses its own IDTM, while an IOS contains tens of thousands of points. In this situation, the network needs to estimate such a large number of IDTMs simultaneously without access to ground truth labels, thus presenting an exceptionally formidable optimization problem. 
Compared with other tasks, teeth possess unique anatomical structures including fixed spatial arrangements of dental categories and local geometric consistency between neighboring points, which can better assist in accurate segmentation.

To solve these challenges, we propose GeoT, a geometry-guided framework that leverages IDTM for semi-supervised tooth segmentation. Specifically, to align IDTM estimation with dental geometric characteristics, we exploit the relatively fixed geometric distribution of tooth categories and design class-level geometric smoothing (CLGS). It encodes the anatomical relationships between tooth categories defined in the FDI notation~\cite{international1984dentistry} (e.g., a molar adjoins premolars and other molars), and constructs a Gaussian prior over transition probabilities, ensuring IDTMs respect these topological constraints. To address the optimization complexity arising from excessive number of IDTMs, we propose point-level geometric regularization (PLGR) to smooth the solution space of IDTM. It minimizes the norm of transition matrix differences between neighboring points of the same category. In this way, IDTM varies gradually within a small neighborhood, ensuring the entire space maintains Lipschitz continuity. This smoothness property significantly simplifies the optimization landscape by reducing the number of local minima and creating more gradual transitions between solutions.

Our main contributions can be summarized as follows: (1) We present GeoT, a practical semi-supervised tooth segmentation framework that learns from noisy pseudo labels. Instead of discarding low-confidence pseudo labels, we use IDTM to model label noise distribution and utilize all noisy labels, representing the first utilization of IDTM for handling pseudo label noise in semi-supervised point cloud segmentation. (2) We leverage both point-level and class-level geometric priors through point-level geometric regularization and class-level geometric smoothing to guide the optimization process of IDTM, significantly enhancing the estimation of IDTM. (3) With extensive experiments on the public Teeth3DS dataset and private dataset, our method achieves state-of-the-art results, attaining performance comparable to fully supervised methods trained on the entire labeled dataset with only $20\%$ of the labeled data.

\section{Method}
\begin{figure}[t]
	\centering
	\includegraphics[scale=0.8]{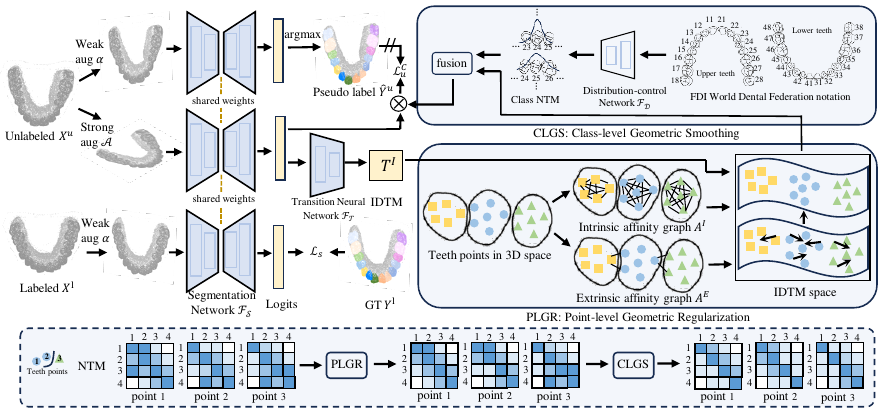}
	\caption{Overview of the proposed GeoT framework. The top portion presents the flowchart of the method, which encompasses the proposed point-level geometric regularization and class-level geometric smoothing. The lower portion illustrates the impact of these two modules on the noise transition matrix. Point 1 and 2 are the same class, point 3 is different. The $4\times4$ grid represents NTM with 4 classes, where a darker color indicates a higher transition probability. Categories 1 and 2, 2 and 3, as well as 3 and 4 are spatially positioned close to each other. PLGR aligns NTMs of the same class and separates different ones. CLGS corrects NTMs violating distribution law. }
	\label{fig:method}
\end{figure}

An overview of our proposed method is illustrated in \figurename~\ref{fig:method}. For labeled data, the network's segmentation results are directly supervised by the ground truth labels. For unlabeled data, we apply both weak and strong data augmentations. The segmentation results from weakly augmented data serve as pseudo-labels. For strongly augmented data, we estimate the IDTM of the prediction results to rectify the supervision signal from the noisy pseudo-labels. To enhance IDTM estimation, we develop point-level geometric regularization and class-level geometric smoothing. PLGR imposes constraints on IDTM by utilizing the geometric priors of dental points in 3D space, while CLGS samples the transfer relationships among categories based on the spatial distribution of tooth classes and incorporates this information into the predicted IDTM. More details will be presented in the following sections.

\subsection{Preliminary}
Formally, the training dataset $\mathcal{D}$ comprises $L$ labeled data and $M$ unlabeled data ($L \ll M$), which can be expressed as $\mathcal{D} = \mathcal{D}^{l} \cup \mathcal{D}^{u}$. Let $C$ and $N$ denote the number of classes and points of each point cloud data, respectively, then $\mathcal{D}^{l} = \{ (X_{i}^{l} \in \mathbb{R}^{N \times 3}, Y_{i}^{l} \in \mathbb{R}^{N \times C}) \}_{i=1}^{L}$ and $\mathcal{D}^{u} = \{ (X_{i}^{u} \in \mathbb{R}^{N \times 3}) \}_{i=L+1}^{L+M}$. Given a labeled data pair $(X^{l}, Y^{l})$\footnote{For the sake of convenience, we omit the subscripts in the subsequent discussion.} and an unlabeled data $X^{u}$, following~\cite{sohn2020fixmatch}, we first perform weak and strong augmentations on them, denoted as $\alpha(\cdot)$ and $\mathcal{A}(\cdot)$, respectively. Then the prediction outputs can be computed by the segmentation network $\mathcal{F_{S}}(\cdot)$, i.e., $P^{l} = \mathcal{F_{S}}(\alpha(X^{l}))$, $P^{u} = \mathcal{F_{S}}(\mathcal{A}(X^{u}))$, and $Q^{u} = \mathcal{F_{S}}(\alpha(X^{u}))$. Through $argmax$ operation on $Q^{u}$, pseudo label $\hat{Y}^{u}$ is determined. Finally, the objective function is the combination of supervised loss $\mathcal{L}_{s}$ and unsupervised loss $\mathcal{L}_{u}$ as 
\begin{equation}
	\mathcal{L} = \mathcal{L}_{s} + \mathcal{L}_{u} = \displaystyle\sum_{k=1}^{N}\mathcal{H}(p^{l}_{k}, y_{k}^{l}) +  \displaystyle\sum_{k=1}^{N}\mathcal{H}(p^{u}_{k}, \hat{y}^{u}_{k}),
    \label{eq:1}
\end{equation}
where $\mathcal{H}(\cdot)$ is the commonly used cross-entropy loss or focal loss~\cite{lin2017focal}, $p^{l/u}_{k}$ is the k-th element of $P^{l/u}$, and $y_{k}^{l}$ ($\hat{y}^{u}_{k}$) is the k-th element of $Y^{l}$ ($\hat{Y}^{u}$). 

\subsection{Noise Transition Matrix for semi-supervised Segmentation}
Since $\hat{Y}^{u}$ is noisy, direct optimizing Eq.~\eqref{eq:1} may lead to model degradation as the segmentation network $\mathcal{F_{S}}(\cdot)$ becomes susceptible to label noise.
To this end, we model the noise in pseudo labels through instance-dependent transition matrix (IDTM) $T^{I}_{k} \in [0, 1]^{C \times C}$, which is leveraged to correct the objective function $\mathcal{L}_{u}$ for robust training. Note that $k\in [1,N]$ indicates the noise transition matrix of k-th point $x_{k}^{u}$. Given that $T^{I}_{k}$ relies on $x_{k}^{u}$, we introduce a transition neural network $\mathcal{F_{T}}(\cdot)$ (e.g., one fully-connected layer) to estimate the transition matrix via $T^{I}_{k} = \mathcal{F_{T}}(p^{u}_{k})$, where $p^{u}_{k}$ denotes the predicted probability distribution over classes for point $x{k}^{u}$. $T^{I}_{k}$ specifies the point-level probability of clean label $m$ being transformed into noisy label $n$ by $T^{I}_{k}[m, n] = p(\bar{Y}=n|Y=m, X=x_{k}^{u})$. Thus, the probability of $x_{k}^{u}$ being classified as $\bar{Y}=n$ is calculated via $p(\bar{Y}=n) = \textstyle\sum_{m=1}^{C} p(Y=m) \cdot T^{I}_{k}[m, n]$, where $p(Y)$ is the clean class distribution. Then the modeled label noise distribution is utilized to rectify the supervision signal (i.e., $\mathcal{L}_{u}$) obtained from noisy pseudo labels. Consequently, the objective function is refined to 
\begin{equation}
	\mathcal{L} = \mathcal{L}_{s} + \mathcal{L}_{u}^{C} = \displaystyle\sum_{k=1}^{N}\mathcal{H}(p^{l}_{k}, y_{k}^{l}) + \displaystyle\sum_{k=1}^{N}\mathcal{H}(p^{u}_{k} T^{I}_{k}, \hat{y}^{u}_{k}).
    \label{eq:2}
\end{equation}
During training, the predicted clean distributions $p^{u}_{k}$ are transformed through the estimated IDTM $T^{I}_{k}$ into noisy distributions that can be directly compared with the noisy pseudo-labels $\hat{y}^{u}_{k}$. This process allows the output of $\mathcal{F_{S}}(\cdot)$ to gradually converge toward clean label distributions, even when only noisy supervision is available. Once properly trained, $\mathcal{F_{S}}(\cdot)$ directly outputs clean class distributions without requiring any further correction by transition matrices at inference time.

 
\subsection{Point-level Geometric Regularization}
Estimating IDTM of a large number of points without appropriate constraints is an ill-posed problem. We propose point-level geometric regularization to addresses this challenge by introducing a geometric prior that constrains the solution space based on a fundamental observation: adjacent points belonging to the same tooth category should exhibit similar noise transition patterns due to their similar local contexts. We encode this prior knowledge by constructing two types of affinity graphs in 3D space: (1) an intrinsic graph connecting each point to its $k_1$-nearest neighbors within the same category, and (2) an extrinsic graph connecting each point to its $k_2$-nearest neighbors from different categories:
\begin{equation}
	A_{ij}^{I} = 	\left\{
	\begin{array}{l}
		e^{-\frac{\left \| x_{i}^{u} - x_{j}^{u}  \right \|^{2}}{\sigma^{2}}},  \ \ if \ \ x_{j}^{u} \in \mathcal{N}(x_{i}^{u}, k_{1}) \ \ and \ \ \hat{y}^{u}_{i} = \hat{y}^{u}_{j},\\
		\quad \quad \quad 0, \qquad \qquad \qquad \qquad  else,
	\end{array} \right.
\end{equation}
\begin{equation}
	A_{ij}^{E} = 	\left\{
	\begin{array}{l}
		e^{-\frac{\left \| x_{i}^{u} - x_{j}^{u}  \right \|^{2}}{\sigma^{2}}},  \ \ if \ \ x_{j}^{u} \in \mathcal{N}(x_{i}^{u}, k_{2}) \ \ and \ \ \hat{y}^{u}_{i} \neq \hat{y}^{u}_{j},\\
		\quad \quad \quad 0, \qquad \qquad \qquad \qquad  else,
	\end{array} \right.
\end{equation}
where $\mathcal{N}(x_{i}^{u}, k_{1/2})$ denotes the $k_{1/2}$-nearest neighbors, $\hat{y}^{u}_{i} = \hat{y}^{u}_{j}$ indicates that $x_{i}^{u}$ and $x_{i}^{u}$ belong to the same tooth, otherwise not. $A_{ij}^{I/E}$ is the intrinsic/extrinsic affinity graph matrix weighted by Gaussian kernel distance. These structural relationships in 3D space are then used to guide the relationships between the corresponding transition matrices as
\begin{equation}
	\mathcal{M}_{I} = \displaystyle\sum_{i,j=1}^{N} A_{ij}^{I} \left \| T^{I}_{i} - T^{I}_{j}  \right \|^{2}, \
    \mathcal{M}_{E} = \displaystyle\sum_{i,j=1}^{N} A_{ij}^{E} \left \| T^{I}_{i} - T^{I}_{j}  \right \|^{2}, 
\end{equation}
\begin{equation}
	\mathcal{L}_{m} = \mathcal{M}_{I} - \mathcal{M}_{E}.
\end{equation}
Here, $\mathcal{M}_{I/E}$ models the relationship between noise transition matrices based on intrinsic/extrinsic affinity graphs in 3D space. Obviously, minimizing $\mathcal{L}_{m}$ encourages the learned IDTM to exhibit proximity if their corresponding points are close in the same class, otherwise be distant from each other. PLGR smooths the IDTM space in a manner that aligns with the physical structure of teeth, reducing the complexity of the optimization problem.

As we can see in the bottom of \figurename~\ref{fig:method}, after PLGR, NTMs of the adjacent points in 3D space of one category (i.e., point 1 and 2) are closer. This is evident in the upper left portions of the NTMs for points 1 and 2, which exhibit increased similarity after PLGR compared to their prior states. Concurrently, PLGR effectively guides the NTMs of points belonging to different classes (i.e., points 2 and 3) to become more distant. As shown in the lower right sections of the NTMs for points 2 and 3, the implementation of PLGR enhances the separation between the NTMs of point 2 and point 3, while simultaneously drawing point 2's NTM closer to that of point 1. This mechanism effectively ensures that the structure within the IDTM space remains consistent with that in the geometric space, thereby mitigating the complexity of the optimization process.

\subsection{Class-level Geometric Smoothing}
Due to the high degree of freedom in the IDTM solution space, it is insufficient to impose soft constraints solely through PLGR. Following FDI World Dental Federation notation in the upper right corner of \figurename~\ref{fig:method}, we can observe that each tooth category is associated with its own specific neighbors. This implies that points within each category exhibit a significantly high probability of transition to some specific categories associated with that category, rather than to other categories. Consequently, we further employ the fixed geometric distribution among tooth categories to guide the optimization of IDTM. Specifically, we leverage Gaussian distributions to model the probability distribution law of transition from the current class to other classes, and utilize a distribution-control network $\mathcal{F_{D}}(\cdot)$ to generate the standard deviation $\sigma$ of the Gaussian distribution for each class, i.e., $\sigma_{m} = \mathcal{F_{D}}(c_{m})$. Here, $c_{m}$ is the m-th class, and $1 \leq m \leq C$. Then we can compute the probability distribution of transition for the m-th category:
\begin{equation}
	P_{m}(c_{n}) = \frac{1}{\sqrt{2\pi}\sigma_{m}} \ e^{-\frac{(c_{n} - c_{m})^{2}}{2\sigma_{m}^{2}}}.
\end{equation}
After calculating each class's distribution, we can sample from them to construct a noise transition matrix $T^{C}$ that incorporates the geometric prior distribution of the categories via $T^{C}[m,n] = P_{m}(c_{n})$. Through a fusion operation on $T^{I}_{k}$ and $T^{C}$, we inject the prior at class level into the transition matrix at point level:
\begin{equation}
	T^{F}_{k} = (1-\lambda)T^{I}_{k} + \lambda T^{C},
\end{equation}
where $\lambda$ is the weighting factor. 
The fusion helps to alleviate noise in the point-level estimation by incorporating robust class-level geometric priors.

As illustrated in the bottom of \figurename~\ref{fig:method}, NTMs that violate the class distribution law (i.e., point 3) are effectively rectified after CLGS. More precisely, we can examine the fourth row of point 3's NTM prior to CLGS, which indicates the transition probabilities of category 4. Notably, it shows a higher likelihood of transitioning to category 2 instead of category 3, despite the spatial proximity of categories 3 and 4, thereby contradicting the expected distribution pattern of tooth classes. In this scenario, CLGS modifies the NTM, and it becomes evident that, post-application of CLGS, point 3's NTM aligns appropriately with the distribution patterns among the categories.

Given that $T^{F}_{k}$ incorporates different levels of geometric priors from teeth, it represents an improved estimation of IDTM, and thus can be utilized to replace $T^{I}_{k}$ in $\mathcal{L}_{u}^{C}$ in Eq.~\eqref{eq:2}. Finally, the overall loss function can be formulated as
\begin{equation}
	\mathcal{L} = \mathcal{L}_{s} + \alpha \mathcal{L}_{u}^{C} + \beta \mathcal{L}_{m} = \displaystyle\sum_{k=1}^{N}\mathcal{H}(p^{l}_{k}, y_{k}^{l}) + \alpha \displaystyle\sum_{k=1}^{N}\mathcal{H}(p^{u}_{k} T^{F}_{k}, \hat{y}^{u}_{k}) + \beta (\mathcal{M}_{I} - \mathcal{M}_{E}),
\end{equation}
where $\alpha$ and $\beta$ are weights of $\mathcal{L}_{u}^{C}$ and $\mathcal{L}_{m}$, respectively. 

\section{Experiments and Results}
\subsection{Dataset and Implementation Details}
We evaluated the proposed method on the public Teeth3DS dataset~\cite{ben2022teeth3ds} and our private dataset. Teeth3DS dataset contains 1800 labeled IOS data with 34 classes collected from 900 patients. Following official split, the dataset was divided into 1200 and 600 scans as training and test sets. We randomly selected 0.5$\%$(6), 1$\%$(12), 5$\%$(60), 10$\%$(120), and 20$\%$(240) of training set as labeled data. Our private dataset comprises 1,200 instances of IOS data collected from 600 patients, which were used as unlabeled data to further validate the effectiveness of the proposed method.

We set $\alpha$, $\beta$, and $\lambda$ to 1.0, 0.1, and 0.9, respectively. $\sigma$ of Gaussian kernel distance in PLGR was 1.0. PointTransformer~\cite{zhao2021point} was adopted as the segmentation network and the transition neural network consisted of one fully-connected layer. The distribution-control network only contained learnable parameters $\sigma_{m}$. Focal loss was used as $\mathcal{H}(\cdot)$ for $\mathcal{L}_{s}$ and $\mathcal{L}_{u}^{C}$. We implemented our method with PyTorch on Nvidia 4090. The model was trained using Adamw optimizer with a learning rate of 1e-3 for 220 epochs and then 1e-4 for another 30 epochs. Following ~\cite{liu2022hierarchical}, we randomly sampled 16,000 points per scan as the input and used a knn-based voting mechanism to upsample the output to original size for better efficiency. Three evaluation metrics were used, including mIoU, Dice Similarity Coefficient (DSC), and point-wise classification accuracy(Acc).

\begin{table}[!h] 
	\centering
	\caption{Comparison of our GeoT with different semi-supervised methods on the Teeth3DS Dataset (\%).}
	\renewcommand\arraystretch{1.2}
    \scriptsize
	\setlength{\tabcolsep}{1.0mm}{
		\begin{tabular}{c|c|c|c|c|c|c|c|c|c|c}
			\hline
			\multicolumn{2}{c|}{ }  &\multicolumn{3}{c|}{Maxillary}&\multicolumn{3}{c|}{Mandible}&\multicolumn{3}{c}{All}\\
			\hline
			Ratio & Method& mIoU& DSC& Acc& mIoU& DSC& Acc& mIoU& DSC& Acc\\
			\hline
            100\%& Upper Bound & $85.27$ & $90.63$& $93.41$& $83.34$ & $88.84$& $90.89$& $84.31$ & $89.74$& $92.15$ \\
            \hline
            \multirow{8}{*}{0.5\%}& SupOnly & $53.65$ & $66.40$& $77.89$& $41.77$ & $54.77$& $66.07$& $47.71$ & $60.59$& $71.98$ \\
			& UA-MT\cite{yu2019uncertainty} &  $57.06$ & $69.21$& $79.94$& $46.64$ & $59.61$& $69.63$& $51.85$ & $64.41$& $74.78$ \\
			& FixMatch\cite{sohn2020fixmatch} & $59.41$ & $71.24$& $79.98$& $45.24$ & $58.07$& $68.82$& $52.33$ & $64.45$& $74.40$\\
			& SLC-Net\cite{liu2022semi} & $58.11$ & $69.97$& $79.53$& $48.49$ & $61.36$& $69.46$& $53.30$ & $65.66$& $74.49$\\
            & DMD\cite{xie2023deep} & $59.38$ & $71.18$& $80.08$& $48.81$ & $61.62$& $70.04$& $54.09$ & $66.40$& $75.06$ \\
            & DC-Net\cite{chen2023decoupled} & $58.06$ & $70.04$& $80.26$& $46.94$ & $59.92$& $69.11$& $52.50$ & $64.98$& $74.69$\\
            & BCP\cite{bai2023bidirectional} & $58.66$ & $70.45$& $79.53$& $48.60$ & $62.19$& $69.83$& $53.63$ & $66.32$& $74.68$\\
            & Ours (GeoT) & $\bm{61.47}$ & $\bm{72.81}$& $\bm{81.02}$& $\bm{52.15}$ & $\bm{64.72}$& $\bm{72.28}$& $\bm{56.81}$ & $\bm{68.76}$& $\bm{76.65}$\\
			\hline
            \multirow{8}{*}{1\%}& SupOnly & $56.30$& $68.20$ & $79.27$& $55.08$& $67.35$ & $74.31$& $55.69$& $67.78$ & $76.79$ \\
			& UA-MT\cite{yu2019uncertainty} &  $61.80$& $72.64$ & $81.72$& $58.62$& $70.16$ & $76.04$& $60.21$& $71.40$ & $78.88$ \\
			& FixMatch\cite{sohn2020fixmatch} & $63.86$& $74.51$ & $82.77$& $60.60$& $71.47$ & $78.55$& $62.23$& $72.99$ & $80.66$\\
			& SLC-Net\cite{liu2022semi} & $62.66$& $73.25$& $82.62$& $61.19$& $72.19$& $78.07$&  $61.93$& $72.72$& $80.34$\\
            & DMD\cite{xie2023deep} & $61.88$& $72.66$& $81.70$& $59.34$& $70.62$& $76.88$&  $60.61$& $71.64$& $79.29$ \\
            & DC-Net\cite{chen2023decoupled} & $61.20$& $71.63$& $81.39$& $59.14$& $70.25$& $76.33$& $60.17$& $70.94$& $78.86$\\
            & BCP\cite{bai2023bidirectional} & $64.02$& $74.27$& $83.13$& $62.06$& $72.74$& $78.53$& $63.04$& $73.51$& $80.83$\\
            & Ours (GeoT) & $\bm{66.90}$& $\bm{76.31}$& $\bm{84.47}$& $\bm{64.07}$& $\bm{74.10}$& $\bm{79.98}$& $\bm{65.48}$& $\bm{75.21}$& $\bm{82.23}$\\
			\hline
            \multirow{8}{*}{5\%}& SupOnly & $72.76$& $81.18$& $87.95$&  $69.76$& $78.82$& $83.94$& $71.26$& $80.00$& $85.95$ \\
			& UA-MT\cite{yu2019uncertainty} &  $77.79$& $84.97$& $90.15$& $72.77$& $80.61$& $85.27$ & $75.28$& $82.79$& $87.71$ \\
			& FixMatch\cite{sohn2020fixmatch} & $75.43$& $83.01$& $89.06$& $72.57$& $80.83$& $85.21$& $74.00$& $81.92$& $87.13$\\
			& SLC-Net\cite{liu2022semi} & $77.02$& $84.53$& $89.78$& $72.16$& $80.41$& $85.04$&  $74.59$& $82.47$& $87.41$\\
            & DMD\cite{xie2023deep} & $77.81$& $85.07$& $90.23$&  $73.44$& $81.42$& $85.71$&  $75.62$& $83.25$& $87.97$\\
            & DC-Net\cite{chen2023decoupled} & $76.90$& $84.21$& $89.73$& $71.67$& $79.58$& $84.91$& $74.29$& $81.89$& $87.32$\\
            & BCP\cite{bai2023bidirectional} & $78.15$& $85.17$& $90.38$& $73.37$& $81.06$& $85.82$& $75.76$& $83.11$& $88.09$\\
            & Ours (GeoT) & $\bm{80.55}$& $\bm{86.96}$& $\bm{91.36}$& $\bm{77.43}$& $\bm{84.38}$& $\bm{87.69}$& $\bm{78.99}$& $\bm{85.67}$& $\bm{89.52}$\\
			\hline
            \multirow{8}{*}{10\%}& SupOnly & $78.11$& $85.36$& $90.34$&  $75.58$& $83.38$& $86.76$& $76.85$& $84.37$& $88.55$ \\
			& UA-MT\cite{yu2019uncertainty} &  $79.38$& $85.96$& $90.87$& $77.41$& $84.43$& $87.81$ & $78.40$& $85.20$& $89.34$ \\
			& FixMatch\cite{sohn2020fixmatch} & $80.17$& $86.77$& $91.25$& $78.21$& $85.31$& $88.27$&  $79.19$& $86.04$& $89.76$\\
			& SLC-Net\cite{liu2022semi} & $79.92$& $86.73$& $91.04$& $77.35$& $84.65$& $87.80$&  $78.64$& $85.69$& $89.42$\\
            & DMD\cite{xie2023deep} & $81.10$& $87.48$& $91.52$&  $78.85$& $85.64$& $88.29$&  $79.98$& $86.56$& $89.91$\\
            & DC-Net\cite{chen2023decoupled} & $80.53$& $86.91$& $91.38$& $78.20$& $85.03$& $88.26$& $79.36$& $85.97$& $89.82$\\
            & BCP\cite{bai2023bidirectional} & $80.04$& $86.79$& $91.11$& $77.63$& $84.87$& $87.96$& $79.83$& $86.33$& $89.93$\\
            & Ours (GeoT) &  $\bm{82.37}$& $\bm{88.18}$& $\bm{92.04}$& $\bm{80.98}$& $\bm{86.97}$& $\bm{89.51}$& $\bm{81.68}$& $\bm{87.57}$& $\bm{90.77}$\\
			\hline
            \multirow{8}{*}{20\%}& SupOnly & $79.57$ & $85.22$& $88.11$& $77.47$& $83.35$ & $85.45$& $78.53$& $84.28$& $86.78$ \\
			& UA-MT\cite{yu2019uncertainty} & $81.22$ & $86.67$& $89.54$& $79.10$& $84.74$ & $86.78$& $80.16$& $85.70$& $88.16$ \\
			& FixMatch\cite{sohn2020fixmatch} & $81.98$ & $87.42$& $90.25$& $80.11$& $85.65$ & $87.65$& $81.04$& $86.53$& $88.95$ \\
			& SLC-Net\cite{liu2022semi} & $80.91$ & $86.35$& $89.21$& $79.60$& $85.09$& $86.91$& $80.25$& $85.72$& $88.06$ \\
            & DMD\cite{xie2023deep} & $83.09$ & $88.52$& $91.44$& $81.90$& $87.37$& $89.18$& $82.49$& $87.95$& $90.31$ \\
            & DC-Net\cite{chen2023decoupled} & $82.94$ & $88.39$& $91.24$& $81.20$& $86.77$& $88.73$& $82.07$& $87.58$& $89.98$ \\
            & BCP\cite{bai2023bidirectional} & $82.86$ & $88.30$& $91.18$& $81.61$& $87.08$& $88.86$& $82.24$& $87.69$& $90.02$ \\
            & Ours (GeoT) & $\bm{84.59}$ & $\bm{90.19}$& $\bm{93.11}$& $\bm{83.08}$& $\bm{88.83}$& $\bm{90.69}$& $\bm{83.83}$& $\bm{89.51}$& $\bm{91.90}$\\
			\hline
            
	\end{tabular}}
	\label{tab:result}
\end{table}

\subsection{Evaluations}
\tablename~\ref{tab:result} compares the quantitative results of our proposed GeoT with different semi-supervised methods. For a fair comparison, we reimplemented these methods using the same backbone. Among these methods, UA-MT~\cite{yu2019uncertainty} and FixMatch~\cite{sohn2020fixmatch} failed to achieve satisfactory results, especially in scenarios with extremely limited labeled data (e.g., 0.5\%). This is because they filtered pseudo labels, resulting in insufficient utilization of unlabeled data. BCP~\cite{bai2023bidirectional} and DMD~\cite{xie2023deep} enriched the supervision signals from their respective perspectives, but they are still suffering from noisy pseudo labels. Hence, the robustness of both methods are weakened. As we can see, DMD only obtained 60.61 mIoU with 1\% labeled data and BCP performed suboptimally on 0.5\% split. Compared to all the above methods, our approach consistently achieves significant improvements across all metrics in all settings, demonstrating the effectiveness of the proposed method. 
 
\begin{table}[t] 
	\centering
	\caption{Comparison of our GeoT with different tooth-specific methods on the Teeth3DS Dataset (\%).}
	\renewcommand\arraystretch{1.2}
    \scriptsize
	\setlength{\tabcolsep}{0.5mm}{
		\begin{tabular}{c|c|c|c|c|c|c|c|c|c|c|c}
			\hline
			\multicolumn{3}{c|}{ }  &\multicolumn{3}{c|}{Maxillary}&\multicolumn{3}{c|}{Mandible}&\multicolumn{3}{c}{All}\\
			\hline
			Ratio & \multicolumn{2}{c|}{Method}& mIoU& DSC& Acc& mIoU& DSC& Acc& mIoU& DSC& Acc\\
			\hline 
            \multirow{5}{*}{0.5\%}&\multirow{2}{*}{MeshSegNet\cite{lian2020deep}} &+ FixMatch\cite{sohn2020fixmatch} & $60.10$ & $70.61$& $79.96$& $49.30$ & $62.35$& $70.54$& $54.70$ & $66.48$& $75.25$ \\
			&  &+ BCP\cite{bai2023bidirectional} &  $60.16$ & $71.14$& $80.01$& $49.66$ & $62.68$& $70.79$& $54.91$ & $66.91$& $75.40$ \\
			& \multirow{2}{*}{TSegNet\cite{cui2021tsegnet}} &+ FixMatch\cite{sohn2020fixmatch} & $60.56$ & $71.01$& $80.32$& $50.02$ & $63.47$& $71.06$& $55.29$ & $67.24$& $75.69$\\
            &  &+ BCP\cite{bai2023bidirectional} & $60.08$ & $71.21$& $80.63$& $48.98$ & $62.01$& $70.23$& $54.53$ & $66.61$& $75.43$\\
            \cline{2-3}
            & \multicolumn{2}{c|}{Ours (GeoT)} & $\bm{61.47}$ & $\bm{72.81}$& $\bm{81.02}$& $\bm{52.15}$ & $\bm{64.72}$& $\bm{72.28}$& $\bm{56.81}$ & $\bm{68.76}$& $\bm{76.65}$\\
			\hline
            \multirow{5}{*}{1\%}& \multirow{2}{*}{MeshSegNet\cite{lian2020deep}} &+ FixMatch\cite{sohn2020fixmatch} & $65.88$& $74.58$ & $83.22$& $61.46$& $73.48$ & $78.98$& $63.67$& $74.03$ & $81.10$ \\
			& &+ BCP\cite{bai2023bidirectional} &  $64.91$& $74.87$ & $83.65$& $62.75$& $73.25$ & $78.87$& $63.83$& $74.06$ & $81.26$ \\
			& \multirow{2}{*}{TSegNet\cite{cui2021tsegnet}} &+ FixMatch\cite{sohn2020fixmatch} & $66.01$& $75.77$ & $83.22$& $62.33$& $72.93$ & $79.36$& $64.17$& $74.35$ & $81.29$\\
            &  &+ BCP\cite{bai2023bidirectional} & $64.64$& $74.49$& $83.50$& $62.12$& $73.01$& $78.40$& $63.38$& $73.75$& $80.95$\\
            \cline{2-3}
            & \multicolumn{2}{c|}{Ours (GeoT)} & $\bm{66.90}$& $\bm{76.31}$& $\bm{84.47}$& $\bm{64.07}$& $\bm{74.10}$& $\bm{79.98}$& $\bm{65.48}$& $\bm{75.21}$& $\bm{82.23}$\\
			\hline
            \multirow{5}{*}{5\%}& \multirow{2}{*}{MeshSegNet\cite{lian2020deep}} &+ FixMatch\cite{sohn2020fixmatch} & $77.89$& $85.11$& $90.15$&  $75.21$& $82.97$& $86.33$& $76.55$& $84.04$& $88.24$ \\
			&  &+ BCP\cite{bai2023bidirectional} &  $79.19$& $86.05$& $89.54$& $74.35$& $82.19$& $87.04$ & $76.77$& $84.12$& $88.29$ \\
			& \multirow{2}{*}{TSegNet\cite{cui2021tsegnet}} &+ FixMatch\cite{sohn2020fixmatch} & $78.88$& $85.66$& $90.62$& $76.02$& $83.44$& $86.78$& $77.45$& $84.55$& $88.70$\\
            & &+ BCP\cite{bai2023bidirectional} & $78.76$& $85.54$& $90.29$& $73.76$& $81.88$& $86.03$& $76.26$& $83.71$& $88.16$\\
            \cline{2-3}
            &\multicolumn{2}{c|}{Ours (GeoT)} & $\bm{80.55}$& $\bm{86.96}$& $\bm{91.36}$& $\bm{77.43}$& $\bm{84.38}$& $\bm{87.69}$& $\bm{78.99}$& $\bm{85.67}$& $\bm{89.52}$\\
			\hline
            \multirow{5}{*}{10\%}& \multirow{2}{*}{MeshSegNet\cite{lian2020deep}} &+ FixMatch\cite{sohn2020fixmatch} & $81.03$& $87.09$& $91.40$&  $79.17$& $85.75$& $88.42$& $80.10$& $86.42$& $89.91$ \\
			&  &+ BCP\cite{bai2023bidirectional} &  $81.82$& $88.03$& $91.70$& $79.46$& $86.11$& $88.62$ & $80.64$& $87.07$& $90.16$ \\
			& \multirow{2}{*}{TSegNet\cite{cui2021tsegnet}} &+ FixMatch\cite{sohn2020fixmatch} & $81.83$& $87.86$& $91.81$& $79.93$& $86.56$& $89.09$&  $80.88$& $87.21$& $90.45$\\
            &  &+ BCP\cite{bai2023bidirectional} & $82.03$& $87.75$& $91.65$& $77.79$& $84.93$& $88.13$& $79.91$& $86.34$& $89.89$\\
            \cline{2-3}
            & \multicolumn{2}{c|}{Ours (GeoT)} &  $\bm{82.37}$& $\bm{88.18}$& $\bm{92.04}$& $\bm{80.98}$& $\bm{86.97}$& $\bm{89.51}$& $\bm{81.68}$& $\bm{87.57}$& $\bm{90.77}$\\
			\hline
            \multirow{5}{*}{20\%}& \multirow{2}{*}{MeshSegNet\cite{lian2020deep}} &+ FixMatch\cite{sohn2020fixmatch} & $83.88$ & $89.11$& $92.23$& $82.06$& $87.45$ & $89.55$& $82.97$& $88.28$& $90.89$ \\
			&  &+ BCP\cite{bai2023bidirectional} & $83.61$ & $89.05$& $92.18$& $82.63$& $88.03$ & $90.08$& $83.12$& $88.54$& $91.13$ \\
			& \multirow{2}{*}{TSegNet\cite{cui2021tsegnet}} &+ FixMatch\cite{sohn2020fixmatch} & $84.31$ & $89.98$& $92.85$& $82.41$& $88.32$ & $90.27$& $83.36$& $89.15$& $91.56$ \\
            &  &+ BCP\cite{bai2023bidirectional} & $82.99$ & $88.46$& $91.49$& $81.87$& $87.28$& $89.35$& $82.43$& $87.87$& $90.4  2$ \\
            \cline{2-3}
            & \multicolumn{2}{c|}{Ours (GeoT)} & $\bm{84.59}$ & $\bm{90.19}$& $\bm{93.11}$& $\bm{83.08}$& $\bm{88.83}$& $\bm{90.69}$& $\bm{83.83}$& $\bm{89.51}$& $\bm{91.90}$\\
			\hline
            
	\end{tabular}}
	\label{tab:result1}
\end{table}

We also compared GeoT with several segmentation methods specifically designed for tooth. Since there are no other semi-supervised tooth-specific methods, we chose fully-supervised tooth segmentation networks \cite{lian2020deep,cui2021tsegnet} as backbones and combined them with semi-supervised algorithms \cite{sohn2020fixmatch,bai2023bidirectional}. The results are presented in \tablename~\ref{tab:result1}. Note that GeoT still used PointTransformer~\cite{zhao2021point} as its backbone, rather than tooth-specific methods. Nevertheless, GeoT achieved SOTA performance, outperforming other methods in all cases, particularly when labeled data is only $0.5\%$ and $1\%$, leading by at least $2.7\%$ and $2.0\%$ in the mIoU metric, respectively. This demonstrates the strong superiority of our approach.

To further explore the potential of GeoT, we utilized the complete training set of the Teeth3DS dataset as labeled data, while employing the data from our private dataset as unlabeled data for experimentation. The results are presented in \tablename~\ref{tab:result2}. It is clear that, compared to other semi-supervised methods, GeoT maximizes improvements in results due to its effective utilization of unlabeled data. Specifically, When employing 600 instances of unlabeled data, GeoT achieves an increase of $2.9\%$ in mIoU, while the use of 1,200 instances leads to an enhancement of $4.6\%$. These findings suggest that GeoT possesses cross-dataset generalizability, enabling it to leverage unlabeled data to significantly improve segmentation outcomes.
 
\begin{figure}[!htp]
	\centering
	\includegraphics[scale=0.42]{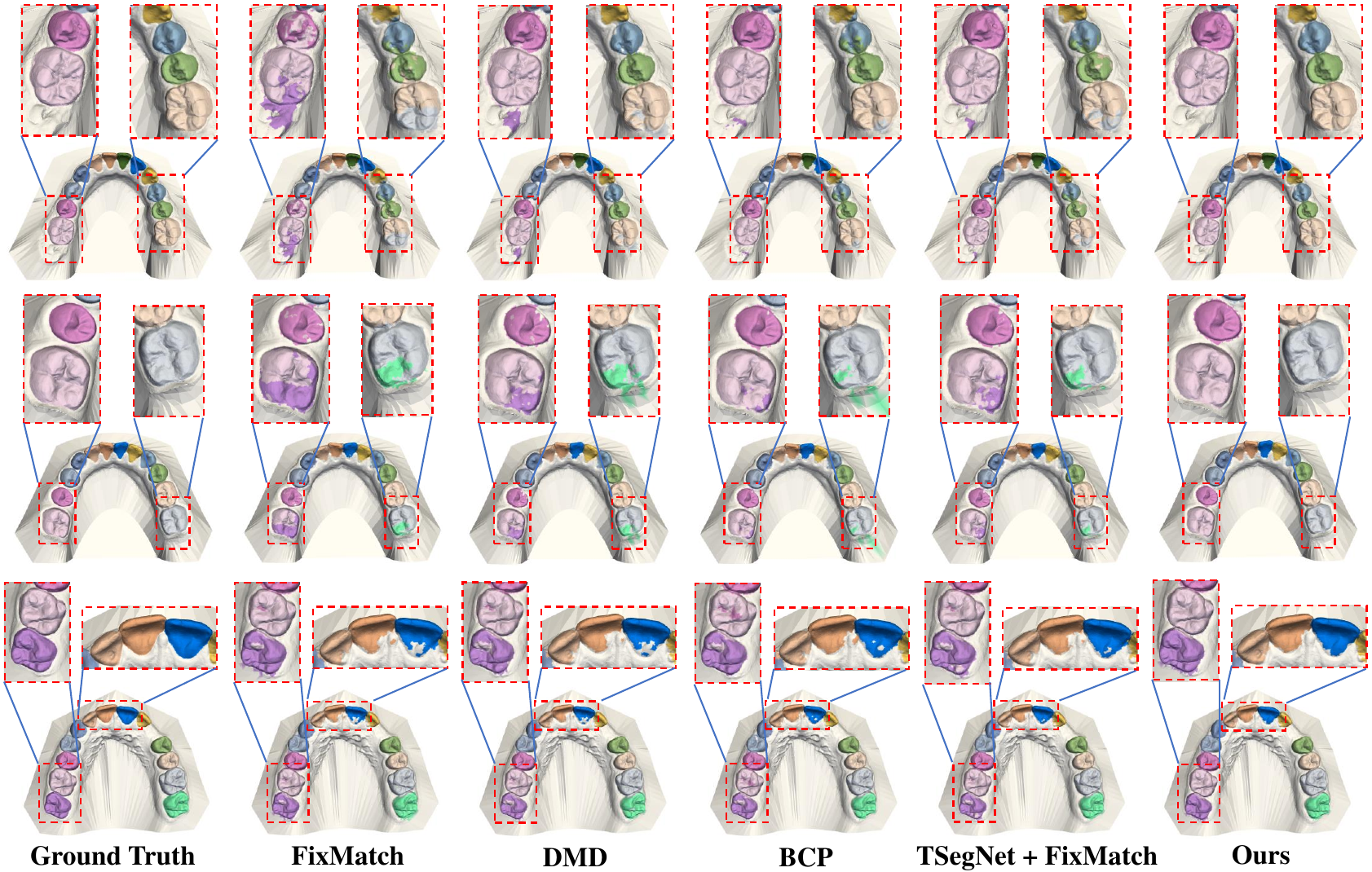}
	\caption{Visualization of the segmentation results of different methods. The first four rows are the lower jaws, and the last two rows are the upper jaws.}
	\label{fig:result}
\end{figure}

Qualitative results with 10\% labeled data are displayed in \figurename~\ref{fig:result} to demonstrate the superior performance of our model. As observed, most methods tend to confuse in adjacent categories during segmentation, especially in the boundary regions of teeth. In contrast, our approach utilizes NTM to handle pseudo-label noise, resulting in clearer supervision signals and achieving better results.

\begin{table}[t] 
	\centering
	\caption{Comparison of our GeoT with different semi-supervised methods on our private dataset and Teeth3DS Dataset (\%).}
	\renewcommand\arraystretch{1.2}
    \scriptsize
	\setlength{\tabcolsep}{1.0mm}{
		\begin{tabular}{c|c|c|c|c|c|c|c|c|c|c}
			\hline
			\multicolumn{2}{c|}{ }  &\multicolumn{3}{c|}{Maxillary}&\multicolumn{3}{c|}{Mandible}&\multicolumn{3}{c}{All}\\
			\hline
			Unlabeled & Method& mIoU& DSC& Acc& mIoU& DSC& Acc& mIoU& DSC& Acc\\
			\hline
            0 & Lower Bound & $85.27$ & $90.63$& $93.41$& $83.34$ & $88.84$& $90.89$& $84.31$ & $89.74$& $92.15$ \\
            \hline
			\multirow{7}{*}{600} & UA-MT\cite{yu2019uncertainty} & $86.69$ & $91.28$& $93.70$& $83.23$ & $89.04$& $91.12$& $84.96$ & $90.16$& $92.41$ \\
			& FixMatch\cite{sohn2020fixmatch} &  $87.36$ & $91.91$& $93.87$& $83.40$ & $89.07$& $91.37$& $85.38$ & $90.49$& $92.62$ \\
			& SLC-Net\cite{liu2022semi} & $87.29$ & $91.99$& $93.77$& $83.71$ & $89.25$& $91.57$& $85.50$ & $90.62$& $92.67$\\
            & DMD\cite{xie2023deep} & $87.80$ & $92.57$& $94.20$& $84.48$ & $89.89$& $91.92$& $86.14$ & $91.23$& $93.06$ \\
            & DC-Net\cite{chen2023decoupled} & $87.06$ & $91.90$& $93.87$& $84.36$ & $89.74$& $91.83$& $85.71$ & $90.82$& $92.85$\\
            & BCP\cite{bai2023bidirectional} & $87.46$ & $92.13$& $94.06$& $84.52$ & $90.03$& $91.96$& $85.99$ & $91.08$& $93.01$\\
            & Ours (GeoT) & $\bm{88.35}$ & $\bm{93.12}$& $\bm{94.57}$& $\bm{85.11}$ & $\bm{90.26}$& $\bm{92.25}$& $\bm{86.73}$ & $\bm{91.69}$& $\bm{93.41}$\\
			\hline
			\multirow{7}{*}{1200} & UA-MT\cite{yu2019uncertainty} &  $87.07$& $91.97$ & $94.24$& $83.63$& $89.33$ & $91.00$& $85.35$& $90.65$ & $92.62$ \\
			& FixMatch\cite{sohn2020fixmatch} & $88.87$& $93.26$ & $94.23$& $85.01$& $90.06$ & $91.83$& $86.94$& $91.66$ & $93.03$\\
			& SLC-Net\cite{liu2022semi} & $88.41$& $92.57$& $94.11$& $84.85$& $89.81$& $91.67$&  $86.63$& $91.19$& $92.89$\\
            & DMD\cite{xie2023deep} & $89.52$& $93.54$& $95.13$& $85.94$& $91.04$& $92.29$&  $87.73$& $92.29$& $93.71$ \\
            & DC-Net\cite{chen2023decoupled} & $88.56$& $93.12$& $94.52$& $85.70$& $90.60$& $92.22$& $87.13$& $91.86$& $93.37$\\
            & BCP\cite{bai2023bidirectional} & $88.98$& $93.21$& $94.62$& $86.12$& $91.13$& $92.46$& $87.55$& $92.17$& $93.54$\\
            & Ours (GeoT) & $\bm{89.72}$& $\bm{93.61}$& $\bm{95.35}$& $\bm{86.58}$& $\bm{91.37}$& $\bm{92.61}$& $\bm{88.15}$& $\bm{92.49}$& $\bm{93.98}$\\
			\hline
            
	\end{tabular}}
	\label{tab:result2}
\end{table}

\subsection{Ablation Study}
We further conducted ablation studies on different ratios (from $0.5\%$ to $5\%$) to verify the effectiveness of each component of GeoT and \tablename~\ref{tab:ablation} reports the results. FixMatch is used as the baseline. By modeling pseudo-label noise using IDTM instead of simply discarding low-confidence samples, the model has shown performance improvements across all four settings. PLGR smooths the solution space of IDTM, reducing the optimization difficulty and further enhancing performance. CLGS leverages prior knowledge of the tooth class distribution to reduce the complexity of IDTM, resulting in significant improvements as well. As the amount of labeled data increases, we observe that the influence of PLGR gradually surpasses that of CLGS. This is attributed to the network gradually capturing the distribution law of the classes under the guidance of labels. Combining all the three components, GeoT demonstrates the most powerful ability for semi-supervised tooth point cloud segmentation. The experimental results indicate that these proposed modules do contribute to the satisfactory performance of our method.

\begin{table}[t] 
	\centering
	\caption{Ablation study of key components (\%).}
	\renewcommand\arraystretch{1.2}
    \scriptsize
	\setlength{\tabcolsep}{1.0mm}{
		\begin{tabular}{ccc|c|c|c|c|c|c|c|c|c}
			\hline
			\multicolumn{3}{c|}{Components}&\multicolumn{3}{c|}{0.5\%}&\multicolumn{3}{c|}{1\%}&\multicolumn{3}{c}{5\%}\\
			\hline
			IDTM& PLGR& CLGS& mIoU& DSC& Acc& mIoU& DSC& Acc& mIoU& DSC& Acc\\
			\hline
            $\times$&$\times$&$\times$ & $52.33$ & $64.45$& $74.40$& $62.23$& $72.99$ & $80.66$& $74.00$& $81.92$& $87.13$ \\
			$\checkmark$&$\times$&$\times$  & $54.13$ & $66.21$& $75.03$& $63.12$& $73.52$ & $80.80$& $76.04$& $83.68$& $88.17$\\
			$\checkmark$&$\checkmark$&$\times$  & $55.45$ & $67.48$& $75.80$& $63.94$& $74.18$ & $81.32$& $77.75$& $84.81$& $88.84$\\
			$\checkmark$&$\times$&$\checkmark$  & $55.91$ & $67.85$& $76.32$& $64.67$& $74.85$& $81.69$& $77.60$& $84.65$& $88.81$\\
            $\checkmark$&$\checkmark$&$\checkmark$  & $\bm{56.81}$ & $\bm{68.76}$& $\bm{76.65}$& $\bm{65.48}$& $\bm{75.21}$& $\bm{82.23}$& $\bm{78.99}$& $\bm{85.67}$& $\bm{89.52}$\\
			\hline
	\end{tabular}}
	\label{tab:ablation}
\end{table}

\subsection{Discussion}
\subsubsection{Hyperparameters}
\begin{figure}[t]
	\centering
	\includegraphics[scale=0.5]{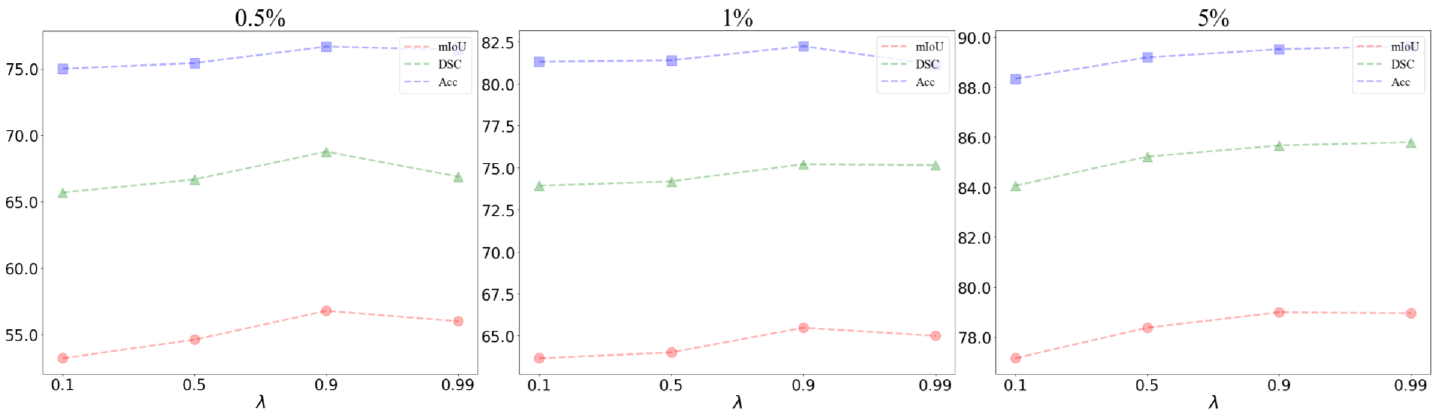}
	\caption{Experiment resuls with different weighting factors $\lambda$ for $T^{F}_{k}$ in different splits.}
	\label{fig:lambda}
\end{figure}

\begin{figure}[t]
	\centering
	\includegraphics[scale=0.5]{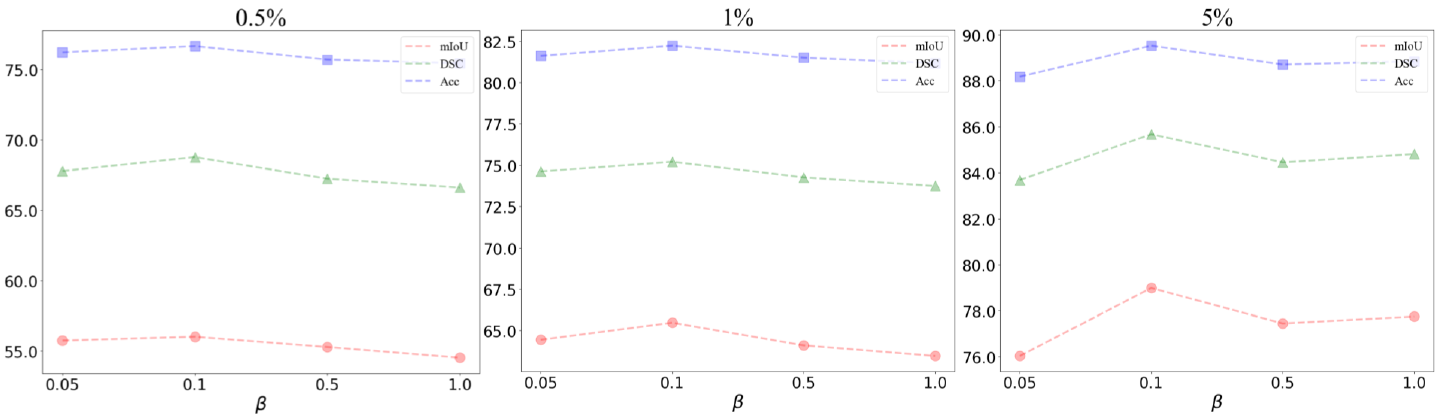}
	\caption{Experiment resuls with different weights $\beta$ for $\mathcal{L}_{m}$ in different splits.}
	\label{fig:beta}
\end{figure}

We investigated the impact of hyperparameters under different partitions (from $0.5\%$ to $5\%$), as shown in \figurename~\ref{fig:lambda} and \figurename~\ref{fig:beta}. \figurename~\ref{fig:lambda} illustrates the effect of the weight factor $\lambda$ on the segmentation results. As $\lambda$ increases (from $0.1$ to $0.9$), more geometric priors are incorporated into the optimization process of IDTM, significantly reducing its estimation difficulty. However, excessive priors ($\lambda = 0.99$) can cause the optimization of IDTM to get stuck in local minima, which affects segmentation accuracy. In \figurename~\ref{fig:beta}, the optimal value of $\beta$ is $0.1$. This is because $\mathcal{L}_{m}$ is designed to maintain the structure of the IDTM space consistent with the 3D space, serving as a spatial regularization term, but it cannot replace the dominant role of the segmentation loss.

\subsubsection{Training Cost of IDTM}
We further computed the training cost of IDTM. Using maxillary teeth with 17 classes as an example, we sample 16,000 points. Each point's NTM is represented as a $17\times17$ matrix, resulting in a total of $16,000 \times 17 \times 17 = 4,624,000$ elements, which is equivalent to a feature size of a $1024\times1024$ image with 4 channels. Therefore, the computational cost is not prohibitive.

\section{Conclusion}
This paper presented a novel framework named GeoT for semi-supervised tooth point cloud segmentation by utilizing instance-dependent transition matrix to handle pseudo label noise. To better estimate the noise transition matrix, on one hand, a regularization module incorporating point-level geometric prior is proposed to regularize the solution space of transition matrices. On the other hand, the prior knowledge of tooth class distribution is utilized to design a smoothing module that guides the optimization of the transition matrix. Extensive experiments showed that our proposed method achieved superior performance on the public dataset, leading to an efficient clinical tool for orthodontic diagnosis.

\subsubsection{\ackname} This work was supported by CUHK 4055188 and SSFCRS 3136048.

%
%
%
%
%
%
\bibliographystyle{splncs04}
\bibliography{paper20}
\end{document}